\documentclass[sigconf]{acmart}

\usepackage{balance}

\def\BibTeX{{\rm B\kern-.05em{\sc i\kern-.025em b}\kern-.08emT\kern-.1667em\lower.7ex\hbox{E}\kern-.125emX}}

\copyrightyear{}
\acmYear{2021}
\setcopyright{acmcopyright}
\acmConference[]{}{}{}
\acmBooktitle{Proceedings of the 30th ACM International Conference on Information and Knowledge Management (CIKM '21), November 1--5, 2021, Virtual Event, Australia}
\acmPrice{15.00}
\acmDOI{00.0000/0000000.0000000}
\acmISBN{000-0-0000-0000-0/00/00}

\usepackage{multirow}
\usepackage{bbding}

\usepackage{amssymb}

\begin{document}

\fancyhead{}

\title{Semantic Borrowing for Generalized Zero-Shot Learning}


\author{Xiaowei Chen}
\affiliation{\institution{Sun Yat-sen University}}

%

%
\begin{abstract}
Generalized zero-shot learning (GZSL) is one of the most realistic
but challenging problems due to the
partiality of the classifier to supervised classes,
especially under the class-inductive instance-inductive (CIII) training
setting, where testing data are not available. Instance-borrowing
methods and synthesizing methods solve it to some
extent with the help of testing semantics, but therefore neither can
be used under CIII. Besides, the latter require
the training process of a classifier after generating examples. In
contrast, a novel non-transductive regularization under CIII called \textbf{Semantic Borrowing (SB)}
for improving GZSL methods with compatibility metric learning is
proposed in this paper, which not only can be used for training linear models,
but also nonlinear ones such as artificial neural networks.
This regularization item in the loss function borrows similar semantics in the training set,
so that the classifier can model the relationship between the
semantics of zero-shot and supervised classes more accurately
during training. In practice, the information of semantics of
unknown classes would not be available for training while this
approach does NOT need it. Extensive experiments on
GZSL benchmark datasets show that SB can reduce the partiality of
the classifier to supervised classes and improve the performance of
generalized zero-shot classification, surpassing inductive GZSL state of the arts.
\end{abstract}

%
%

%

%

%
\maketitle

\section{Introduction}
\label{sec:introduction}

Classification has made great progress driven by the advancement
of deep learning, but a large number of instances for each class are
required, and the classifiers trained on the instances for training
cannot classify instances of the classes that the previous instances
don't belong to. These challenges severely limit the application of
these classification methods in practice. Many methods have been
proposed to overcome these difficulties \cite{2019A}, including zero-shot learning \cite{2009Learning,larochelle2008zero,palatucci2009zero} and generalized zero-shot learning (GZSL) \cite{chao2016empirical,xian2018zero}. The semantic meaning of the
label of a class can be defined by training examples of the class in
traditional classification problems, but different from it, the
semantic meaning of the label of an unseen class cannot be defined by
training examples in GZSL. To solve this problem, a semantic
space can be defined, in which each label of a seen or unseen class
is identified uniquely. There are three training settings for a
GZSL classifier. Class-transductive instance-inductive (CTII) setting
allows the use of testing semantics during training, class-transductive instance-transductive (CTIT) setting also
allows the use of unlabeled testing example features, and class-inductive instance-inductive (CIII)
setting allows neither of these two. Their further descriptions can be found
in \cite{2019A}. The existing GZSL methods can be divided into six groups \cite{2019A}, namely
correspondence, relationship, combination, projection,
instance-borrowing and synthesizing methods. Due to the differences in the distributions of the seen and
unseen classes, a GZSL classifier will suffer from the domain shift
problem \cite{fu2015transductive}, which reduces the accuracy of generalized zero-shot learning \cite{chao2016empirical}.
Instance-borrowing methods \cite{jiang2019adaptive} and synthesizing
methods \cite{zhu2018generative,xian2018feature,sariyildiz2019gradient} solve this problem to some extent with the help of testing semantics, but therefore neither of them can be used under CIII \cite{2019A}
where testing data are invisible, and the latter always require the
training process of a classifier after generating examples based on
testing semantics.

\begin{figure}[!t]
\begin{center}
\centerline{
{
{\includegraphics[width=1.03\columnwidth]{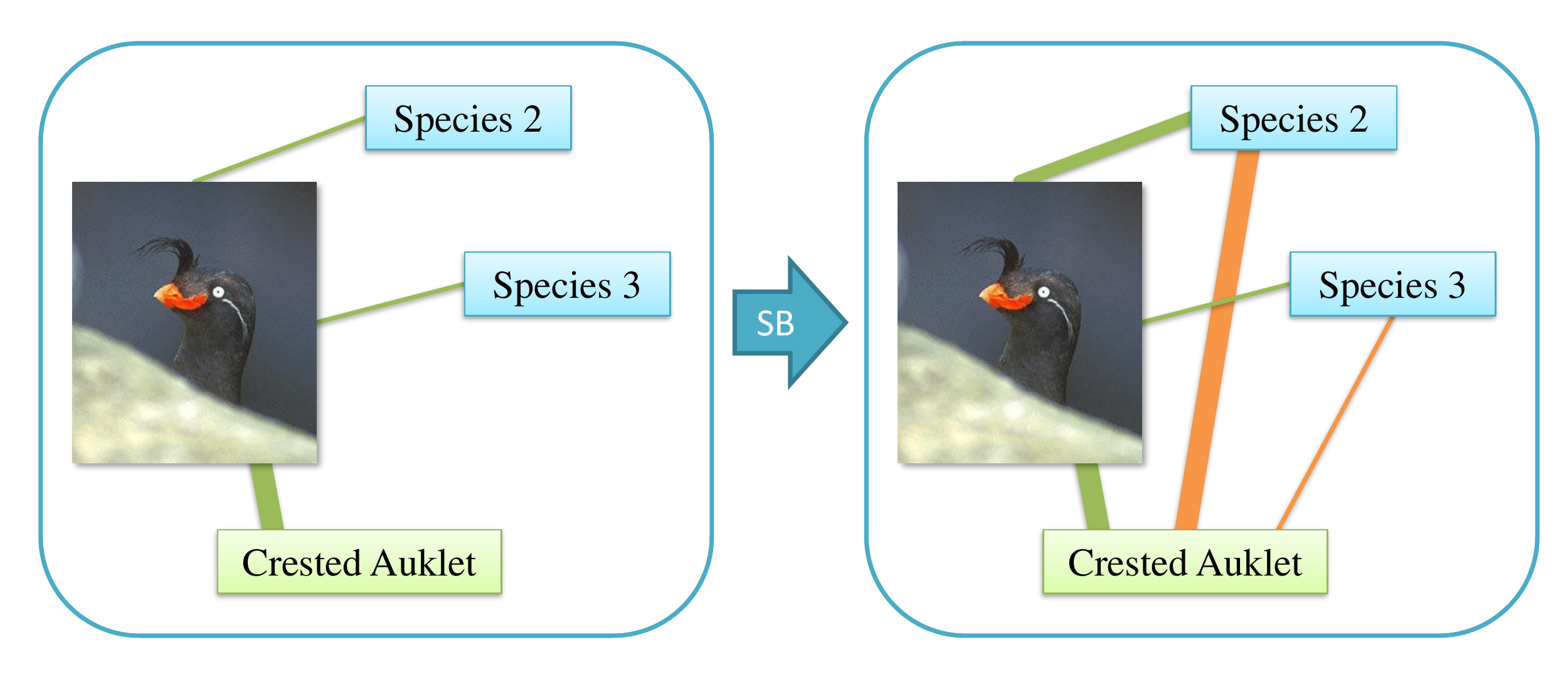}}}
}
\caption{Illustration of the improvement of compatibilities by
SB. The thickness of the line between a photo and text
indicates the compatibility between them while the one
between two texts indicates the semantic similarity
between them. By borrowing similar semantics in the training
set, we can enable a classifier to model more accurately the
relationship between the semantics of unseen and seen classes
during training without semantics of unseen classes.}
\label{fig:Semantic borrowing}
\end{center}\vskip -0.1in
\end{figure}

In this paper, a non-transductive regularization is proposed to improve the
compatibility metric learning used in GZSL methods under CIII. In
the GZSL methods based on compatibility metric learning, the
relationship between features and semantics, that is, compatibility,
is learned through metric learning, and then the differences among
the compatibilities between a testing feature and all semantic
candidates in this metric space are determined, and finally, the
semantic candidate corresponding to the testing example feature is
determined accordingly, so that the class label of the testing feature
can be obtained, thus achieving the goal of GZSL. Different from
the process above, by additionally borrowing similar semantics in the training
set, we can enable a classifier to model the relationship between
the semantics of unseen and seen classes more accurately during
training without the semantics of unseen classes, thereby reducing
the partiality of the classifier to seen classes during testing to deal
with the domain shift problem, as shown in Figure~\ref{fig:Semantic borrowing}. The proposed
regularization is named \textbf{Semantic Borrowing (SB)}.

The main contributions are highlighted as follows: \textbf{1)} In practice, the semantics of unknown classes would not be available for training. So different from instance-borrowing methods and synthesizing ones, this
approach utilizes neither semantics nor instances of unknown classes, totally under the strict but realistic CIII \cite{2019A} training setting. \textbf{2)} As a regularization, this approach not only can be used for training linear models, but also nonlinear ones such as artificial neural networks, improving GZSL methods with compatibility metric learning.

\section{The Proposed Regularization}
\label{sec:proposed method}

SB is applied to the compatibility metric learning in GZSL methods.
As mentioned earlier, GZSL with compatibility metric learning
will learn the compatibilities between features and semantics
through metric learning. At the same time, SB learns additionally
the compatibility between each feature and the most similar
semantic vector to the semantic vector corresponding to the feature.
With the help of this information, the differences among the
compatibilities between a testing feature and all semantic
candidates in the learned metric space will be more accurate. In
other words, the relationship between the semantics of unseen and
seen classes is modeled more accurately by the classifier. SB is
illustrated in the right panel of Figure~\ref{fig:Semantic borrowing}.

The set of all seen classes is denoted as $B_s$
 and the set of all
unseen classes $B_u$, $B_s \cap B_u = \varnothing$, then the set of all classes $B = B_s \cup B_u$.
For any class $b \in B$, there is a unique corresponding
semantic vector $s \in \mathbb{R}^n$. The set of all semantic vectors is denoted
as $S$, and the set of all semantic vectors of seen classes $S_s$, then the
set of all seen-class examples $D_s \subseteq \left\{ (f,s) \mid f \in F_s,s \in S_s \right\}$, where
$F_s \subsetneq F \subseteq \mathbb{R}^m$ is the set of all features of seen-class examples, and
$F$ is the set of all features of examples. The set of all unseen-class
examples is denoted as $D_u$, then
GZSL learns a classifier on the
training set $D_{tr} \subseteq \left\{ (f,s) \mid f \in F_{tr},s \in S_{tr} \right\} \subsetneq D_s$
 to obtain the classes of example features in testing sets $D_{te-s} \subsetneq D_s$
 and $D_u$,
where $D_{tr} \cap D_{te-s} = \varnothing$.

\subsection{Preparing Models for Regularization}

The compatibilities between features and semantics form a metric
space in which the compatibility between a feature and its
corresponding semantic vector will be greater than those between
the feature and other semantics. In order to learn such a space, we
can use a linear model or a nonlinear one to fit it, but because
they have different fitting capabilities due to the different
complexities of a linear model and a nonlinear model, we need
define different objectives to train them.

For the linear model, in order to adapt to its limited fitting
ability, we can train a compatibility function $c:F \times S \rightarrow \mathbb{R}$ on the
training dataset with the objective of symmetric structured joint
embedding in the previous multi-modal structured learning
methods \cite{akata2015label,akata2015evaluation,reed2016learning}:
\begin{small}\begin{align}
\label{eq:formula1}
L_0^{(u)}(f_i,s_i;\theta)=L_f^{(u)}(f_i,s_i;\theta)+L_s^{(u)}(f_i,s_i;\theta),
\end{align}
\end{small}where $(f_i,s_i) \in D_{tr}^{(u)} \subseteq D_{tr}$ and the two misclassification losses are:
\begin{small}\begin{align}
\label{eq:formula2}
L_f^{(u)}(f_i,s_i;\theta)=\frac{\sum\limits_{s \in S_{tr}^{(u)} \setminus \{ s_i \}} \max \{ 0,1+c(f_i,s;\theta)-c(f_i,s_i;\theta) \}}{|S_{tr}^{(u)}|-1},
\end{align}
\begin{align}
\label{eq:formula3}
L_s^{(u)}(f_i,s_i;\theta)=\frac{\sum\limits_{f \in F_{tr}^{(u)} \setminus \{ f_i \}} \max \{ 0,1+c(f,s_i;\theta)-c(f_i,s_i;\theta) \}}{|F_{tr}^{(u)}|-1},
\end{align}
\end{small}where $S_{tr}^{(u)} \subseteq S_{tr}$, $F_{tr}^{(u)} \subseteq F_{tr}$, $| \cdot |$ indicates the cardinality of a set, $B \setminus A$ denotes the relative
complement of $A$ in $B$.

For the nonlinear model, because of its strong fitting ability,
we can use the MSE loss to train a compatibility function on the
training set as in \cite{sung2018learning}. Therefore, $L_0^{(u)}$ in Eq. (\ref{eq:formula1}) becomes:
\begin{small}\begin{align}
\label{eq:formula6}
L_0^{(u)}(f_i,s_i;\theta)=\frac{\sum\limits_{s \in S_{tr}^{(u)} \setminus \{ s_i \}} c^2(f_i,s;\theta)}{|S_{tr}^{(u)}|-1}+[c(f_i,s_i;\theta)-1]^2.
\end{align}\end{small}

\subsection{Semantic Borrowing Regularization}
After preparing the model that will be trained with Semantic
Borrowing (SB), it is time to add SB regularization to its loss function.
In order for the classifier to model the relationship
between the semantics of unseen and seen classes more accurately
during training, SB adds a new objective that borrows similar
semantics in the training set. It is different from instance-borrowing methods, which borrow data in the testing set.

For the linear model above, the SB regularization is:
\begin{small}
\begin{align}
\label{eq:formula4}
L_{SB}^{(v)}(f_i,s_i,s_j;\theta)=\frac{\sum\limits_{s \in S_{tr}^{(v)} \setminus \{ s_j \}} \max \{ 0,1+c(f_i,s;\theta)-c(f_i,s_j;\theta) \}}{|S_{tr}^{(v)}|-1},
\end{align}
\end{small}where $s_j \in S_{tr}^{(v)} \subseteq S_{tr}$ is the most similar semantic vector in the current second
training subset $S_{tr}^{(v)}$ to $s_i$
 in the current first
training subset $S_{tr}^{(u)}$.

For the nonlinear model above, the SB regularization is formulated correspondingly as:
\begin{small}\begin{align}
\label{eq:formula7}
L_{SB}^{(v)}(f_i,s_i,s_j;\theta)=\frac{\sum\limits_{s \in S_{tr}^{(v)} \setminus \{ s_j \}} c^2(f_i,s;\theta)}{|S_{tr}^{(v)}|-1}+[c(f_i,s_j;\theta)-1]^2.
\end{align}\end{small}

Finally, the overall loss for a model trained with SB is:
\begin{small}\begin{align}
\label{eq:formula5}
L^{(t)}(\theta) & = \underbrace{ \sum\limits_{(f_i,s_i) \in D_{tr}^{(2t)}} L_0^{(2t)}(f_i,s_i;\theta) }_{\text{Compatibility metric learning}}  \\ \nonumber
& + \underbrace{ \alpha \sum\limits_{(f_i,s_i) \in D_{tr}^{(2t)} \atop s_j=C^{(2t+1)}(s_i)} L_{SB}^{(2t+1)}(f_i,s_i,s_j;\theta) }_{\text{Semantic borrowing}}
+ \beta \| \theta \|_2,
\end{align}
\end{small}where $t=0,1, 2,\cdots$, $\alpha \in (0,1)$, $\beta$ controls weight decay, \begin{small}$C^{(v)}:S_{tr} \rightarrow S_{tr}^{(v)}$\end{small} is used to
find similar semantics. By minimizing this loss, we can make the
compatibility between a feature and the most semantically similar
semantic vector to it also greater than those between the feature and
other semantics.

\subsection{Semantic Similarities}
When using SB to improve GZSL methods with compatibility
metric learning, it is necessary to borrow the most similar semantic
vector in the training set to each training semantic vector, which
requires the calculation of the semantic similarity. Thanks to the
process of determining the similarity in SB independent of the
objective, in the case that the training semantics are equal-dimensional vectors of attributes,
we can use the negative mean absolute error (-MAE) as the semantic similarity to make the semantic comparison more precise. Compared with the negative mean square error, cosine similarity and Ruzicka similarity \cite{deza2009encyclopedia},
using -MAE can get better results on h and u in experiments.
Therefore, the function for seeking similar semantics can be
formulated as:
\begin{small}\begin{align}
\label{eq:Formula 8}
C^{(v)}(s_i)=\mathop{\operatorname{argmin}}_{s \in S_{tr}^{(v)}}\| s-s_i \|_1.
\end{align}\end{small}

\subsection{Classification}
By minimizing Eq. (\ref{eq:formula5}), we can obtain the compatibilities between
features and semantics. Based on the learned compatibility
function, a multi-class classifier $M:F \rightarrow S$, that achieves the goal of
GZSL can be formulated as follows:
\begin{small}\begin{align}
\label{eq:Formula 9}
M(f)=\mathop{\operatorname{argmax}}_{s \in S}c(f,s),
\end{align}
\end{small}where $f \in F$. Then the class corresponding to $M(f)$ is what we
want.

\section{Experiments}

\subsection{Evaluation \& Implementation}

In order to evaluate SB, CUB \cite{wah2011caltech} and SUN \cite{patterson2012sun}
are selected as the representatives of fine-grained benchmark datasets,
and AWA1 \cite{2009Learning}, AWA2 \cite{xian2018zero} and aPY \cite{farhadi2009describing} as the representatives
of coarse-grained benchmark datasets. The splits, semantics and evaluation metrics used in the
comparison are proposed in \cite{xian2018zero}, where semantics are class-level
attributes. Different from \cite{sariyildiz2019gradient}, no additional semantics are used for CUB.
If the length range of semantic vectors in a dataset is small,
it will be scaled to be consistent with that in the other dataset.
Following \cite{xian2018zero,xian2018feature,bucher2017generating}, example features are the 2048-dimensional top pooling units of a ResNet-101 pretrained on
ImageNet-1K, without any preprocessing.
Average per-class top-1
accuracies in \% (T-1) are calculated as evaluation scores. The
metrics u and s are T-1 of unseen and seen classes, respectively,
and h is their harmonic mean \cite{xian2018zero}. u reflects the performance of a
classifier for unseen classes, s reflects the performance for seen
classes, and h indicates the comprehensive performance.

The experiments comprehensively evaluate SB with different
models. The bilinear mapping \cite{sariyildiz2019gradient} is selected as the representative
of the linear model, and the multilayer perceptron (MLP) combination used in \cite{sung2018learning} as the
representative of the nonlinear model. The combination consists of
two MLPs with one hidden layer, and the numbers of hidden units are
hyperparameters. The first MLP maps semantics into the
feature space, and the second MLP maps the concatenations of features
and mapped semantics into compatibilities. Each layer has a ReLU
activation function, except for the last layer with a sigmoid
activation function. The former model is optimized with minibatch SGD while the latter model is optimized with Adam.

\subsection{Comparison with Inductive GZSL State of the Arts}

There have been methods that can be used to solve the
GZSL problem to some extent. Compared with them, we can see
that SB can build new power for GZSL. In Table~\ref{tab:comparison with some present methods}, linear models
and nonlinear models trained with SB are both compared with state-of-the-art inductive GZSL methods.

\begin{table*}
\caption{Comparison with GZSL state of the arts on the
benchmarks from \cite{xian2018zero}. Their results are taken from the papers. The results of linear models are listed in
the upper half of the table, and the results of nonlinear models in
the lower half. In each half, the methods above \emph{Trained with SB} are counterparts,
and the methods below \textit{Trained with SB} are NOT counterparts.
h reflects the comprehensive performance.}
\centering
\scriptsize
\begin{tabular}{l|ccc|ccc|ccc|ccc|ccc}
\toprule
\multirow{2}{*}{\textbf{Method}} & \multicolumn{3}{c|}{\textbf{CUB}}             & \multicolumn{3}{c|}{\textbf{SUN}}             & \multicolumn{3}{c|}{\textbf{AWA1}}             & \multicolumn{3}{c|}{\textbf{AWA2}}             & \multicolumn{3}{c}{\textbf{aPY}}            \\ \cline{2-16}
                                 & \textbf{u}    & \textbf{s}    & \textbf{h}    & \textbf{u}    & \textbf{s}    & \textbf{h}    & \textbf{u}    & \textbf{s}    & \textbf{h}    & \textbf{u}    & \textbf{s}    & \textbf{h}    & \textbf{u}    & \textbf{s}    & \textbf{h}    \\ \midrule
DAP \cite{2009Learning}           & 1.7           & 67.9          & 3.3           & 4.2           & 25.1          & 7.2           & 0.0           & \textbf{88.7} & 0.0           & 0.0           & 84.7          & 0.0           & 4.8           & 78.3          & 9.0            \\
IAP \cite{2009Learning}                      & 0.2           & \textbf{72.8} & 0.4           & 1.0           & 37.8          & 1.8           & 2.1           & 78.2          & 4.1           & 0.9           & 87.6          & 1.8           & 5.7           & 65.6          & 10.4           \\
CONSE \cite{norouzi2013zero}                   & 1.6           & 72.2          & 3.1           & 6.8           & 39.9          & 11.6           & 0.4           & 88.6          & 0.8           & 0.5           & \textbf{90.6}          & 1.0           & 0.0           & \textbf{91.2}          & 0.0           \\
ALE \cite{akata2013label}                     & 23.7          & 62.8          & 34.4          & {21.8}           & 33.1          & \textbf{26.3}           & 16.8          & 76.1          & 27.5           & 14.0           & 81.8          & 23.9           & 4.6           & 73.7          & 8.7          \\
SYNC \cite{changpinyo2016synthesized}                    & 11.5          & 70.9          & 19.8          & 7.9           & \textbf{43.3}          & 13.4           & 8.9           & 87.3          & 16.2           & 10.0           & 90.5          & 18.0           & 7.4           & 66.3          & 13.3          \\
\hline
\textbf{Trained with SB}                         & \textbf{29.1} & 59.8          & \textbf{39.1} & \textbf{22.8}          & 30.7          & 26.2          & \textbf{21.8} & 86.1          & \textbf{34.8}           & \textbf{17.2}           & 89.2          & \textbf{28.8}           & \textbf{18.2}           & 73.0          & \textbf{29.1} \\ \hline
*AML   \cite{jiang2019adaptive}                      & 25.7           & 66.6          & 37.1          & 20.0           & 38.2          & \textbf{26.3}           & 11.8           & 89.6          & 20.8           & -           & -          & -           & 12.6           & 74.5          & 21.5           \\ \midrule
RN \cite{sung2018learning}             & 38.1          & 61.4          & 47.0          & -           & -          & -           & 31.4          & \textbf{91.3} & 46.7           & 30.0           & \textbf{93.4}          & 45.3           & -           & -          & - \\
DEM \cite{zhang2017learning}           & 19.6          & 57.9          & 29.2          & 20.5           & 34.3          & 25.6           & 32.8          & 84.7 & 47.3           & 30.5           & 86.4          & 45.1           & 11.1           & 75.1          & 19.4          \\
EDEM \cite{zhang2020towards}           & 21.0          & 66.0          & 31.9          & 22.1           & 35.6          & 27.3           & \textbf{36.9}          & 90.6 & \textbf{52.4}           & \textbf{35.2}           & 93.0          & \textbf{51.1}           & 7.8           & 75.3          & 14.1          \\ \hline
\textbf{Trained with SB}                         & \textbf{41.7}          & \textbf{64.2} & \textbf{50.6} & \textbf{23.1}           & \textbf{42.9}          & \textbf{30.0}           & 36.5          & 86.7          & 51.4           & {34.8}           & 89.2          & 50.1           & \textbf{16.1}           & \textbf{86.9}          & \textbf{27.2}          \\ \hline
*GAZSL \cite{zhu2018generative}    & 31.7          & 61.3          & 41.8          & 22.1           & 39.3          & 28.3           & 29.6          & 84.2          & 43.8           & -           & -          & -           & 14.2           & 78.6          & {24.0}          \\
*GMN \cite{sariyildiz2019gradient} & \textbf{56.1} & 54.3 & 55.2 & \textbf{53.2} & 33.0 & 40.7 & 61.1 & 71.3 & 65.8 & - & - & - & - & - & - \\
*EDEM\_ex \cite{zhang2020towards}  & {54.0}          & 62.9          & \textbf{58.1}          & {47.2}           & 38.5          & \textbf{42.4}           & \textbf{71.4}          & 90.1 & \textbf{79.7}           & \textbf{68.4}           & 93.2          & \textbf{78.9}           & \textbf{29.8}           & 79.4          & \textbf{43.3} \\           \bottomrule
\end{tabular}
\label{tab:comparison with some present methods}
\end{table*}

Whether among linear or nonlinear models, it is easy to see that models trained
with SB get the best h and u, except in a few cases, but the scores are still almost equal to the best ones.
It shows that they are less biased towards seen classes than those without SB and the comprehensive
performance is also improved, as described in Section~\ref{sec:introduction}.
It needs to be added that, unlike
all other models in the table, which are trained under the CIII
training setting where testing data are invisible, GAZSL and GMN use
testing semantics to synthesize examples for unseen classes so as
to learn the final classifier, so it is impossible for them to
be used under CIII. Therefore, they are NOT counterparts. AML and EDEM\_ex are NOT, either.
The comparison with all of these is added here
for completeness. In fact, the use of SB in a synthesizing
method with compatibility metric learning can be a future study,
where SB will be used in non-CIII training settings.

\subsection{Effectiveness}

In order to verify the effectiveness of SB, an ablation study is
conducted here. Table~\ref{tab:effectiveness} demonstrates the comparison of models
trained with and without SB. It shows SB improves h and
u of both linear and non-linear models on both fine-grained and
coarse-grained datasets, in some cases also improves s, thanks
to the more accurately modeled relationship between the semantics
of unseen and seen classes with SB.

\begin{table}
\caption{Comparison of models trained with and
without SB.}
\centering
\scriptsize
\begin{tabular}{l|ccc|ccc}
\toprule
\multirow{2}{*}{\textbf{Model}} & \multicolumn{3}{c|}{\textbf{CUB}}             & \multicolumn{3}{c}{\textbf{AWA1}}            \\ \cline{2-7}
                                & \textbf{u}    & \textbf{s}    & \textbf{h}    & \textbf{u}    & \textbf{s}    & \textbf{h}    \\ \midrule
Linear                          & 27.2          & \textbf{59.9} & 37.4          & 18.0          & 84.3          & 29.6          \\
Linear+SB                       & \textbf{29.1} & 59.8          & \textbf{39.1} & \textbf{21.8} & \textbf{86.1} & \textbf{34.8} \\ \midrule
Nonlinear                       & 40.0          & 63.0          & 48.9          & 32.5          & \textbf{87.9} & 47.4          \\
Nonlinear+SB                    & \textbf{41.7} & \textbf{64.2} & \textbf{50.6} & \textbf{36.5} & 86.7          & \textbf{51.4} \\ \bottomrule
\end{tabular}
\label{tab:effectiveness}
\end{table}

\subsection{Effect}

\begin{figure}[!t]
\begin{center}
\centerline{
{
{\includegraphics[width=1.03\columnwidth]{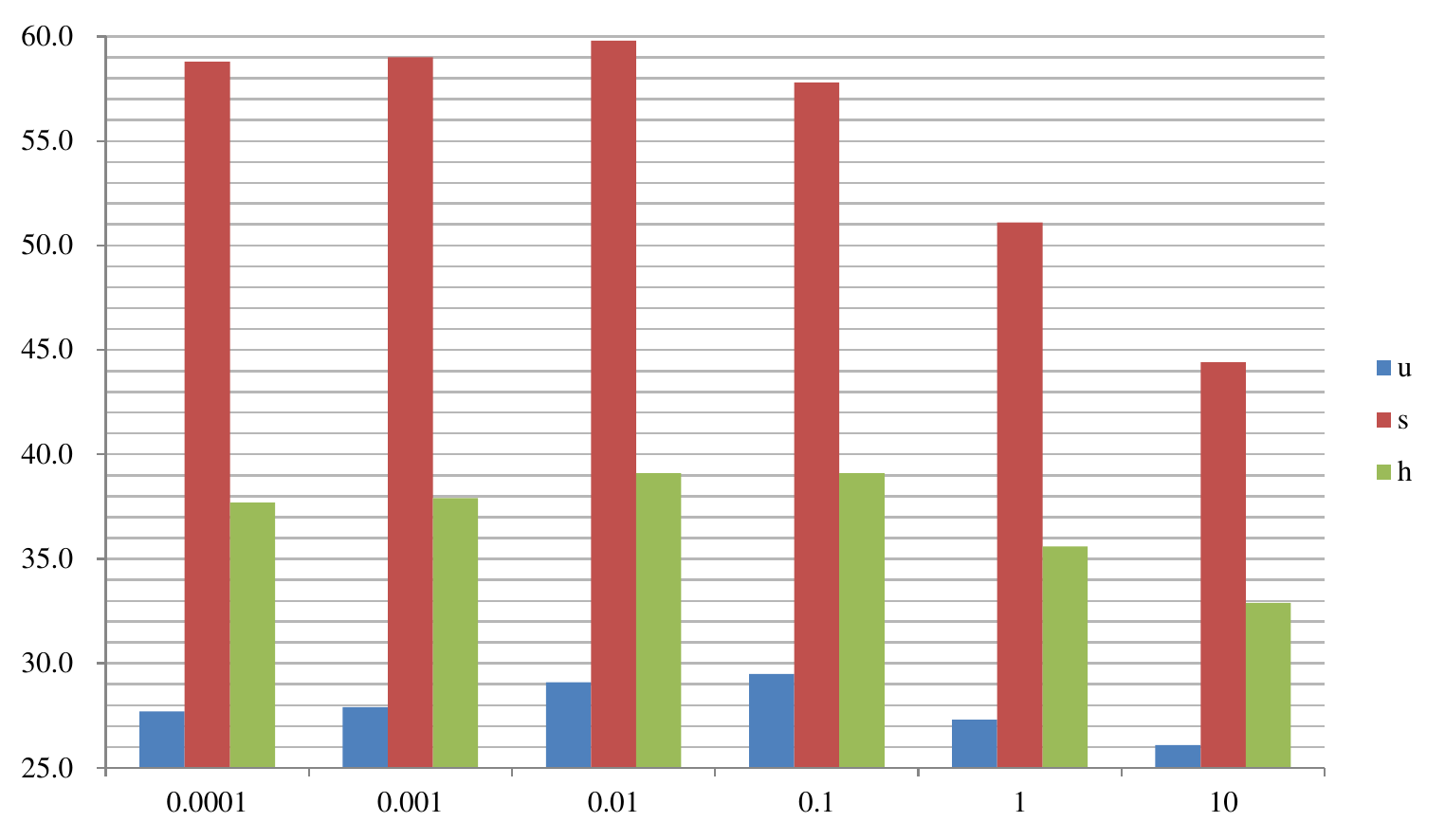}}}
}
\caption{Analysis of the influence of $\alpha$ on u, s and h scores of a linear model trained with SB on CUB.}
\label{fig:Effect}
\end{center}\vskip -0.1in
\end{figure}

The effect of SB on the original method is affected by $\alpha$ in Eq. (\ref{eq:formula5}).
By evaluating models trained with different $\alpha$, the way SB takes effect can be more clear.
For this, a set of linear
models are trained with different $\alpha$ on CUB. Figure~\ref{fig:Effect} shows the
evaluation results of six representative values of $\alpha$. Combined with
Table~\ref{tab:effectiveness}, it can be seen that the models are worse than those trained
without SB when $\alpha \geqslant 1$. It is expected because the compatibility
between each feature and its semantically similar semantic vector
is learned additionally with SB, so that the relationship between the
semantics of unseen and seen classes is modeled more accurately,
which improves the performance of the GZSL classifier, but when
each compatibility of this kind is greater than or equal to the
compatibility between the feature and its corresponding semantic
vector, the relationship modeling becomes worse. In addition, we
can observe that the model obtains the best h and s when $\alpha=0.01$
and the best h and u when $\alpha=0.1$. On the both sides, the
performance of the model decreases. It shows again that modeling
a too large or too small compatibility between each feature and its
semantically similar semantic vector will lead to inaccuracy,
thereby reducing the improvement of the original method by SB.

\section{Conclusion}
In this work, non-transductive semantic borrowing regularization is proposed to
improve GZSL methods with compatibility metric learning under
CIII. Extensive evaluation of representative models trained on
representative GZSL benchmark datasets with the proposed
regularization has shown that it can improve the performance of
generalized zero-shot classification, surpassing inductive GZSL state of the arts.


\bibliographystyle{ACM-Reference-Format}
\balance
\bibliography{CIKM-sigconf}

\end{document}